\documentclass[10pt,journal,compsoc]{IEEEtran}
\usepackage{graphicx}
\usepackage{amssymb}
\usepackage{mdframed}
\usepackage{lineno}
\usepackage{amsmath}
\usepackage{subcaption}

\ifCLASSOPTIONcompsoc
  \usepackage[compress]{cite}
\else
  \usepackage{cite}
\fi
\ifCLASSINFOpdf
\else
\fi
\begin{document}
%
\title{Markov Random Field Model-Based \\ Salt and Pepper Noise Removal}
%
%
%
%

\author{Ahmadreza~Baghaie
}


\IEEEtitleabstractindextext{%
\begin{abstract}
Problem of impulse noise reduction is a very well studied problem in image processing community and many different approaches have been proposed to tackle this problem. In the current work, the problem of fixed value impulse noise (salt and pepper) removal from images is investigated by use of a Markov Random Field (MRF) models with smoothness priors. After the formulation of the problem as an inpainting problem, graph cuts with $\alpha$-expansion moves are considered for minimization of the energy functional. As for comparisons, several other minimization techniques that are widely used for MRF models' optimization are considered and the results are compared using Peak-Signal-to-Noise-Ratio (PSNR) and Structural Similarity Index (SSIM) as metrics. The investigations show the superiority of graph cuts with $\alpha$-expansion moves over the other techniques both in terms of PSNR and also computational times. 
\end{abstract}

}

\maketitle

\IEEEdisplaynontitleabstractindextext

\IEEEpeerreviewmaketitle

\IEEEraisesectionheading{\section{Introduction}\label{sec:introduction}}

\IEEEPARstart{N}{oise}  reduction as one of the first pre-processing steps in image processing has gained immense attention throughout the years and there exists a plethora of literature introducing novel approaches to deal with various types of noises that are commonplace in digital image processing and the related fields. Among different types of noise, Gaussian noise, speckle noise and impulse noise can be mentioned. Gaussian noise, which is independent from pixel intensity, is mainly caused by Johnson-Nyquist noise (thermal noise) \cite{ohta2007smart,nakamura2005image}. Speckle noise, on the other hand, is common in images acquired by narrow-band detection systems like Synthetic Aperture Radar (SAR), ultrasound and Optical Coherence Tomography (OCT) systems and its distribution can be represented by a Rayleigh distribution \cite{baghaie2015sparse,baghaie2016application,baghaie2015state}. On the other hand, impulse noise is usually caused during the process of acquisition and transmission of digital images \cite{gonzalez2002digital,bovik2010handbook}. There are two types of impulse noise: random-valued (RV) impulse noise and fixed-valued (FV) impulse noise, which is also known as \textit{salt and pepper}. In the RV, the corrupted pixels are replaced with a random value while in the case of FV the corrupted pixels take values equal to minimum or maximum possible intensity levels. The general practice for FV noise removal mainly involves some forms of mean filtering, median filtering, compressed sensing or image inpainting \cite{chan2005salt, chen2008efficient, toh2008salt, toh2010noise, huang2010removal, dharmarajan2010hypergraph, jayaraj2010new, wu2011efficient, esakkirajan2011removal, sree2013salt, zhang2014new, sahin2014salt, ahmed2014removal, li2014modified, guo2014salt, gao2015efficient, li2015salt, sun2015efficient}. The process generally starts with detecting the corrupted pixels in the image which is later followed by the attempt for estimating the true values from the known pixels. 

In \cite{chan2005salt}, at first, an adaptive median filter is used for detecting the noisy pixels. By minimizing a regularized energy functional that is only applied to the noisy pixels, the true image is restored. Chen et al. \cite{chen2008efficient} use a directional correlation-dependent filtering technique for recovering the noisy pixels. Use of fuzzy logic has also been explored for the process of salt and pepper noise removal from images \cite{sahin2014salt,ahmed2014removal}. Toh et al. proposed to use a fixed \cite{toh2008salt} and adaptive \cite{toh2010noise} fuzzy switching median filtering approach for detection and correction of noisy pixels. In \cite{sahin2014salt} a two dimensional cellular automata (CA) combined with fuzzy logic is used for removal of salt and pepper noise. Ahmed and Das \cite{ahmed2014removal} use an adaptive fuzzy detector followed by a weighted mean filtering approach for suppression of high density salt and pepper noise. Compressed sensing and sparse representation based techniques have also proven to be useful for removal of salt and pepper noise. In \cite{huang2010removal} the noisy pixels are first detected by a simple min/max detection in the range of possible intensity values. Then by assuming that the true image is compressible, the dual-tree complex wavelet transform (DT-CWT) is used as the sparse representation matrix which is later used for reconstruction of the noise-free image. Guo et al. \cite{guo2014salt} used a similar approach by considering a path-based non-local operator in combination with adaptive median filter for detecting noisy pixels. A very recent example of sparse representation-based techniques can be found in the work of Sun et al. \cite{sun2015efficient} in which, a modified boundary discriminative noise detection approach and iterative shearlet transform are used for detection and removal of salt and pepper noise, respectively. 

Even though not exhaustive, the presented list of literature for solving the problem of salt and pepper impulse noise in digital image processing is a clear representation of various techniques proposed for solving the problem of salt and pepper noise removal. However, to the best of our knowledge, use of probabilistic graphical models and Markov Random Field (MRF) modeling has never been explored for solving this problem. MRF modeling and the range of its applications in image processing and computer vision has been studied intensively in the literature \cite{szeliski2010computer, li2012markov}, with applications in stereo vision, optical flow estimation, image registration, image restoration etc \cite{so2011non, szeliski2008comparative, sun2003stereo, baghaie2015dense, baghaie2016dense}. In the current work, the problem of FV noise removal from images are formulated as an inpainting MRF modeling problem with smoothness priors. Then using graph cuts with expansion moves, the energy functional is minimized which results in a reconstructed version of the corrupted image. Several well-known test images with various levels of salt and pepper noise, ranging from 10 to 90\%, are considered. Moreover, different techniques for minimizing the energy functional are considered for completeness of the comparisons using PSNR and SSIM as metrics. 

The paper is organized as follows: Section \ref{S:2} starts with a brief introduction of fixed-valued impulse noise model, and then moves to a more elaborate discussion on the general MRF modeling of computer vision and image processing problems as well as graph cuts with $\alpha$-expansion moves. The section is concluded by the proposed method which discusses the specific choices made here for solving the problem of salt and pepper noise removal. This is followed in Section \ref{S:3} by introducing the comparison metrics as well as results of the proposed method and a few state-of-the art techniques for minimizing the energy functional defined for the problem. This section is concluded by some notes on the possible improvements in the future as well as the option of using the same concept for super-resolution of images in order to increase the resolution. Finally, Section \ref{S:4} concludes the paper. 

\section{Methods}
\label{S:2}
\subsection{Salt and Pepper Noise Model}
The classical salt and pepper impulse noise model can be defined as follows \cite{chan2005salt}. Assume $x(i,j)$ as the pixel located at $(i,j)$ in a $M\times N$ image grid and $[l_{min}, l_{max}]$ as the dynamic range of the intensity values. The noisy gray level of the pixel located at $(i,j)$ is given by
\begin{equation}
x_n(i,j)=
\begin{cases}
l_{min},  \quad \text{with probability $P$} \\
l_{max}, \quad \text{with probability $Q$} \\
x(i,j), \, \, \text{with probability $1-P-Q$}
\end{cases}
\end{equation}
where $R=P+Q$ defines the noise level. In the current work, various levels of salt and pepper noise, ranging from $10-90\%$ are considered. 

\subsection{Graph Cuts with $\alpha$-Expansion Moves for Markov Random Field (MRF) Models with Smoothness Priors}
Considering the pixel labeling problem, the aim is to assign to every pixel $p \in \mathcal{P}$ a label, $l_p \in \mathcal{L}$ \cite{szeliski2008comparative}. In this scenario $\mathcal{P}$ is the set of image pixels, $\mathcal{L}$ represents the collection of possible labels, $n$ is the number of pixels and $m$ is the number of labels. The problem of labeling can be formulated as minimization of the energy functional $E$ as follows:
\begin{equation}
E=E_{data}+\lambda E_{smooth}
\label{eq:general}
\end{equation}
where $E_{data}$ and $E_{smooth}$ are representatives of \textit{data} and \textit{smoothness} terms, respectively and $\lambda$ is for modifying the amount of contribution of each term in the energy functional. In this formulation, $E_{data}$ measures the difference between the labeling and the observed data while $E_{smooth}$ is a representative of the extent to which the labeling is not piecewise smooth \cite{boykov2001fast}. Generally speaking, the data term can be considered as a set of per pixel data costs $D_p(l)$:
\begin{equation}
E_{data}=\sum_p D_p(l_p)
\end{equation}
where $D_p$ measures the degree to which $l_p$ fits pixel $p$ considering the observed data. On the other hand, the smoothness term can be seen as the sum of spatially varying horizontal/vertical nearest neighbor smoothness costs, $V_{pq}(l_p,l_q)$:
\begin{equation}
E_{smooth}=\sum_{\{p,q\} \in \mathcal{N}} V_{pq}(l_p,l_q)
\end{equation}
where $\mathcal{N}$ is the set of all neighboring pixel pairs. 

The graph cuts with $\alpha$-expansion moves as proposed by \cite{boykov2001fast} for minimizing the energy functional represented in Eq.(\ref{eq:general}) can be summarized as follows. Consider the input labeling $l$ (partition $\mathbf{P}$) and a label $\alpha$, the aim is to find a labeling $\hat{l}$ which minimizes the energy functional within one $\alpha$-expansion of $l$ over all labelings. Assuming the interaction penalty function $V$ as a \textit{metric}, it satisfies:

\begin{equation}
\begin{aligned}
(a) \ \quad \ V(\alpha, \beta) =0 \quad \Leftrightarrow \quad \alpha=\beta,\\
(b) \quad \ V(\alpha, \beta) \quad = \quad V(\beta, \alpha)\geq 0,\\
(c) \quad \ V(\alpha, \beta) \leq V(\alpha, \gamma)+V(\gamma, \beta)
\end{aligned}
\end{equation}
far any labels $\alpha, \beta, \gamma \in \mathcal{L}$. Therefore, the process of finding the best expansion move is based on computing a labeling corresponding to a minimum cut on a graph $\mathcal{G}_{\alpha}=<\mathcal{V}_{\alpha},\mathcal{E}_{\alpha}>$ where $\mathcal{V}_{\alpha}$ and $\mathcal{E}_{\alpha}$ represent the nodes and edges of the graph, respectively \cite{boykov2001fast}. It should be noted that the structure of the graph is dictated by the current labeling and by the label $\alpha$. The set of vertices are defined as
\begin{equation}
\mathcal{V}_{\alpha}=\Big\{\alpha, \bar{\alpha}, \mathcal{P}, \bigcup_{\substack{\{p,q\}\in \mathcal{N}, \\ l_p \neq l_q}} a_{\{p,q\}} \Big\}
\end{equation}
where $\alpha$ and $\bar{\alpha}$ are the two terminals, $\mathcal{P}$ is the set of image pixels and $\bigcup a_{\{p,q\}}$ is the set of auxiliary nodes that are placed between neighboring nodes that are separated given the current labeling (partition). The set of edges in graph $\mathcal{G}_{\alpha}$ are

\begin{equation}
\mathcal{G}_{\alpha}=\Big\{ \bigcup_{p\in \mathcal{P}} {\{t_p^\alpha, t_p^{\bar{\alpha}}\}}, \bigcup_{\substack{\{p,q\}\in \mathcal{N}, \\ l_p = l_q}} e_{\{p,q\}}, \bigcup_{\substack{\{p,q\}\in \mathcal{N}, \\ l_p \neq l_q}} \mathcal{E}_{\{p,q\}}  \Big\}
\end{equation}
where $t_p^\alpha$ and $t_p^{\bar{\alpha}}$ are edges (\textit{t}-link) connecting the pixel $p$ to the terminals $\alpha$ and $\bar{\alpha}$, $\bigcup e_{\{p,q\}}$ is the set of edges between neighboring pixels that are not separated by the current labeling and $\bigcup \mathcal{E}_{\{p,q\}}$ is the set of triplet edges between the neighboring pixels separated by the current lableing and the auxiliary node between each pair. Each edge is assigned a specific weight based on the data and smoothness term's values associated with each node. 

Any cut $\mathcal{C}$ on the graph $\mathcal{G}_{\alpha}$ must include only one \textit{t}-link for any pixel. Moreover, the labeling that corresponds to this cut, $l^{\mathcal{C}}$, should be one $\alpha$-expansion away from the initial labeling. It is proven in \cite{boykov2001fast} that for ${p,q}\in \mathcal{N}$ and $l_p\neq l_q$, the minimum cut $\mathcal{C}$ on the graph satisfies $|\mathcal{C}\cap \mathcal{E}_{p,q}|=V(l_p^{\mathcal{C}}, l_q^{\mathcal{C}})$. Also, for any cut $\mathcal{C}$, the cost of the cut can be computed as $|\mathcal{C}|=E(l^{\mathcal{C}})$. These were later shown to provide the basis for the ultimate goal: when expansion moves are allowed, the local minimum will be within a known factor of the global optimum. This makes the graph cuts algorithm among the best methods for minimization of MRF modeling problems with applications in stereo matching, optical flow estimation, image registration, image restoration etc.

\subsection{Proposed Approach}

The first step in most of impulse noise detection techniques is locating the noisy pixels and many sophisticated techniques have been proposed in the literature, especially for random-valued impulse noise \cite{dong2007new,dong2007detection,chen2001adaptive}. Even though not very accurate in extreme cases of having very bright or dark regions in the images, here, the simple min/max detection is used for locating the noisy pixels. Of course, this step can be replaced with any other technique for locating the noisy pixels without having any effects on the following steps.  Having the location of the corrupted pixels, a mask can be generated indicating these pixels as missing pixels from the image that need to be estimated. This way, the problem is considered as an inpainting problem that we need to solve using a MRF-based energy minimization approach. 

Choice of proper data and smoothness cost functions is of high importance in solving various problems. For the problem of image restoration and denoising considered here, the data term for each pixel is defined as the squared difference between the assigned label and the observed intensity. This can be defined as:
\begin{equation}
D_p=(l_p-I_p)^2
\end{equation}
where $I_p$ is the observed intensity. Of course, this is only the case for the observed regions, while for the obscured or missing regions, the data term is zero. Having the squared difference as the data term will penalize big deviations from the observed pixel values which is required for the problem of salt and pepper noise removal.

Depending on the specific problem, the smoothness term can result in an overall smooth or discontinuity preserving labeling. In image denoising, the smoothness term usually is defined as a function of the absolute difference of the adjacent labels:
\begin{equation}
V(|l_p-l_q|)=min(|l_p-l_q|^k, V_{max})
\end{equation}
with $k\in \{1,2\}$ and $V_{max}$ indicates the bound on the largest possible penalty. Depending on the choice of $k$, the result will be an absolute or quadratic truncated distance interaction penalty, which are examples of discontinuity preserving cost functions that allow for having several regions of similar labels resulting in a piecewise smooth labeling. For the current work, $k=2$ is chosen. The value for $V_{max}$ is set to 5 experimentally. Moreover, the standard first order (4-connected) neighborhood system is used for $\mathcal{N}$ which makes the smoothness energy as the sum of spatially varying vertical and horizontal closest neighbor smoothness costs. The value of $\lambda$ is also set to 5 experimentally. Figure \ref{fig:gcAlg} shows the algorithm of the proposed approach.

\begin{figure}[bt]
\centering
\begin{mdframed}
\newcommand{\keyw}[1]{{\bf #1}}
\begin{tabbing}
\quad \=\quad \=\quad \kill
\textbf{Algorithm: Proposed Approach}\\
\noindent\rule{8cm}{0.4pt}\\
\keyw{1.} \textbf{Initialization:}\\
\> $\circ$ $E=E_{data}+\lambda E_{smooth}$;\\
\> $\circ$ $D_p=(l_p-I_p)^2$;\\
\> $\circ$ $V(|l_p-l_q|)=min(|l_p-l_q|^k, V_{max})$\\
\> \quad \quad with $k=2$, $V_{max}=5$, $\lambda=5$;\\
\keyw{2.} \textbf{Noise Detection \& Mask Creation:}\\
\> $\circ$ Min/Max intensity detection; \\
\keyw{3.} \textbf{Graph Cuts with $\alpha$-Expansion:}\\
\> $\circ$ Starting from an arbitrary labeling $l$;\\
\> \keyw{for} each label $\alpha \in \mathcal{L}$ \keyw{do}\\
\> \{ \\
\> \quad \keyw{find} $\hat{l}=argmin \ E(l')$ among $l'$ within \\
\> \> \qquad one $\alpha$-expansion of $l$\\
\>  \quad\keyw{if} $E(l')<E(l)$, set $l:=\hat{l}$\\
\> \} \\
\> \keyw{return} $l$
\end{tabbing}
\end{mdframed}

\caption{Proposed approach for salt and pepper noise removal using MRF modeling with smoothness priors and $\alpha$-expansion based graph cuts }
\label{fig:gcAlg}
\end{figure}

\section{Experimental Results}
\label{S:3}
\subsection{Comparison Metrics}

Assessing the performance of the proposed algorithm based on graph cuts with $\alpha$-expansions moves for the problem of salt and pepper impulse noise removal from images require a proper set of metrics. For this, Peak-Signal-to-Noise-Ratio (PSNR) and Structural SIMilarity index (SSIM) are used. 

Assuming $x$ and $y$ as the original and reconstructed images respectively, each with the size of $M\times N$ pixels, the Mean Squared Error (MSE) can be defined as:
\begin{equation}
MSE=\frac{1}{MN}\sum_{i=1}^M \sum_{j=1}^N [x(i,j)-y(i,j)]^2
\end{equation}

Therefore the PSNR (in dB) can be defined as:
\begin{equation}
PSNR=10.\log_{10} \Big(\frac{L^2}{MSE}\Big)
\end{equation}
where $L$ is the maximum possible pixel value in the image.

SSIM can be defined as follows \cite{wang2004image}:
\begin{equation}
SSIM(x,y)=\frac{(2\mu_x \mu_y+c_1)(2\sigma_{xy}+c_2)}{(\mu_x^2+\mu_y^2+c_1)(\sigma_x^2+\sigma_y^2+c_2)}
\end{equation}
where $\mu$ and $\sigma$ are representative of the mean and variance of the corresponding images and $\sigma_{xy}$ is the covariance of the two images. Assuming the dynamic range of the images to be $L$, $c_1=(k_1L)^2$ and $c_2=(k_2L)^2$ with $k_1=0.01$ and $k_2=0.03$ as the default values. 

\subsection{Results \& Discussions}
\begin{figure*}[hbtp]
\centering
\begin{subfigure}[b]{.19\textwidth}
\centering
\includegraphics[scale=.17]{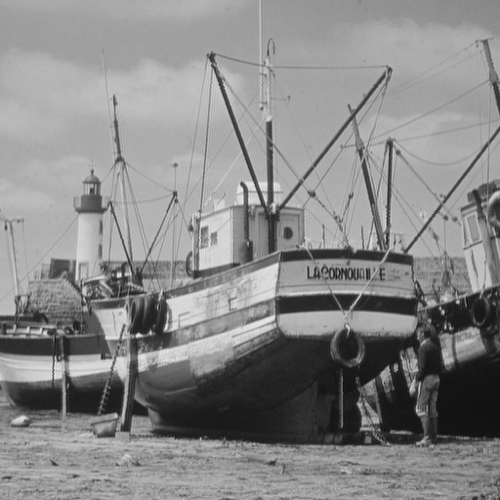}
\caption{}
\end{subfigure}
\begin{subfigure}[b]{.19\textwidth}
\centering
\includegraphics[scale=.17]{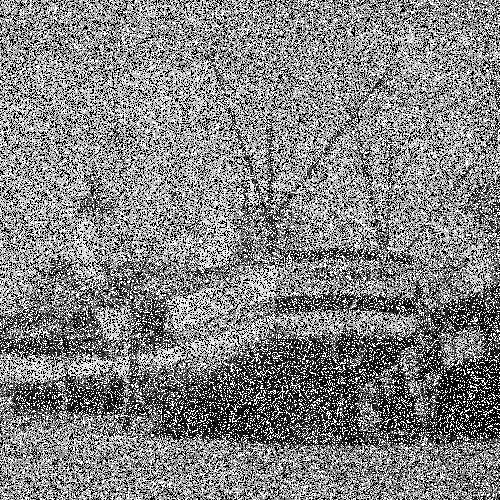}
\caption{}
\end{subfigure}
\begin{subfigure}[b]{.19\textwidth}
\centering
\includegraphics[scale=.17]{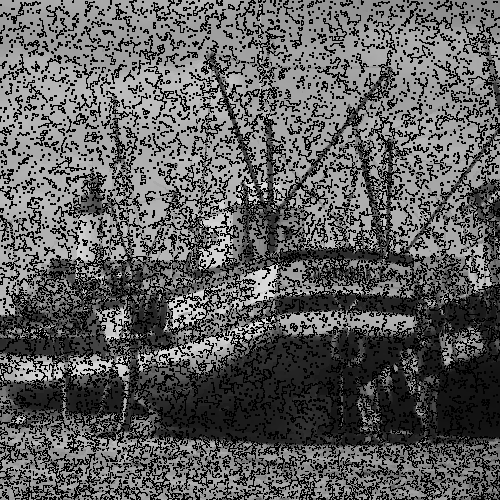}
\caption{}
\end{subfigure}
\begin{subfigure}[b]{.19\textwidth}
\centering
\includegraphics[scale=.17]{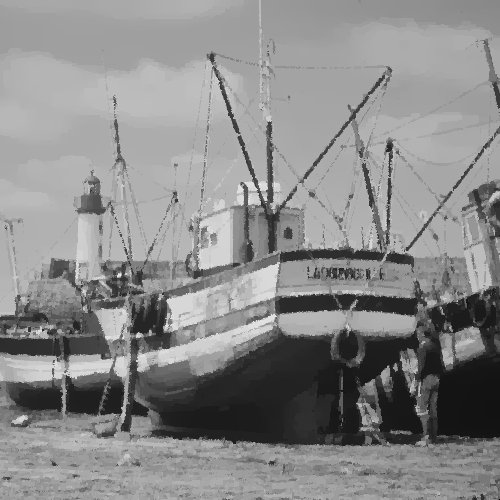}
\caption{}
\end{subfigure}\\
\begin{subfigure}[b]{.19\textwidth}
\centering
\includegraphics[scale=.17]{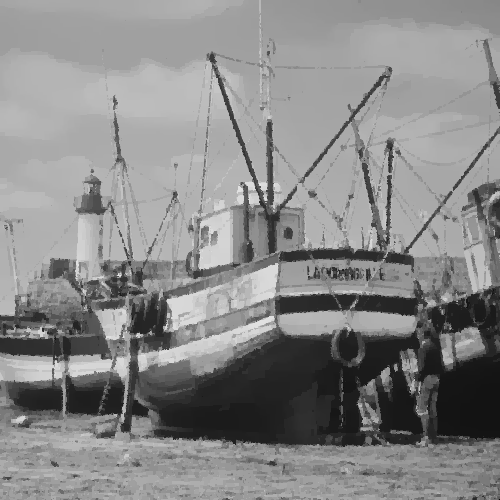}
\caption{}
\end{subfigure}
\begin{subfigure}[b]{.19\textwidth}
\centering
\includegraphics[scale=.17]{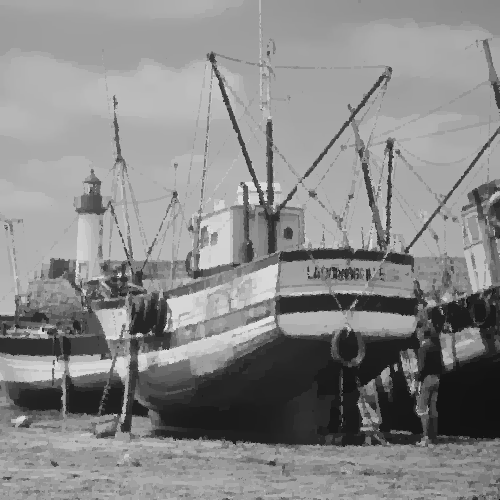}
\caption{}
\end{subfigure}
\begin{subfigure}[b]{.19\textwidth}
\centering
\includegraphics[scale=.17]{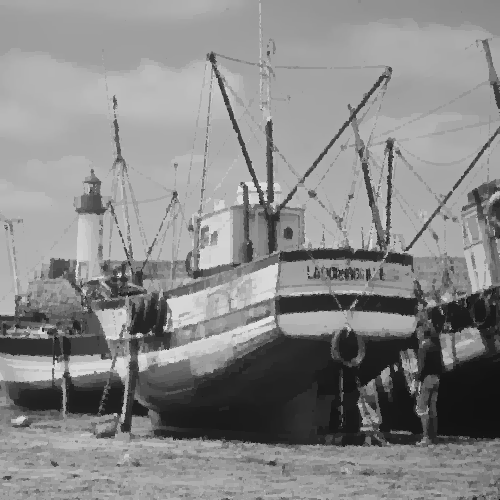}
\caption{}
\end{subfigure}
\begin{subfigure}[b]{.19\textwidth}
\centering
\includegraphics[scale=.17]{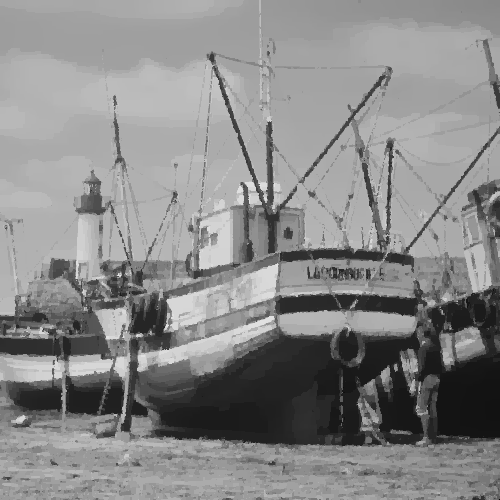}
\caption{}
\end{subfigure}
\vskip -0.35cm
\caption{The results of \textit{Boat} corrupted by 50\% salt and pepper noise, the numbers indicate PSNR and SSIM: (a) Ground truth, (b) Noisy images (8.35, 0.03), (c) ICM (10.05, 0.12), (d) Swap (27.45, 0.86), (e) BP-S (28.08, 0.87), (f) TRW-S (28.08, 0.87), (g) BP-M (28.08, 0.88), (h) Expansion (\textbf{28.38}, 0.88)}
\label{fig:boatResults}
\end{figure*}

\begin{figure*}[hbtp]
\centering
\begin{subfigure}[b]{.19\textwidth}
\centering
\includegraphics[scale=.17]{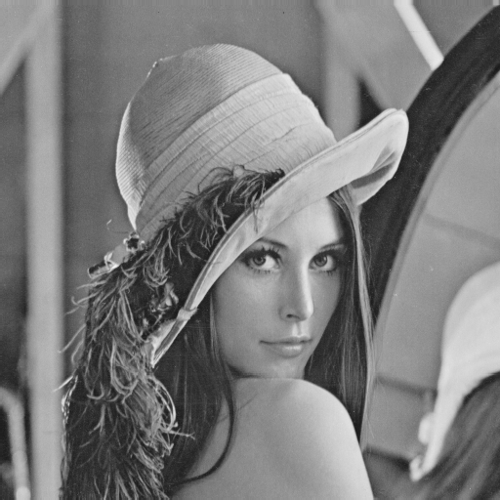}
\caption{}
\end{subfigure}
\begin{subfigure}[b]{.19\textwidth}
\centering
\includegraphics[scale=.17]{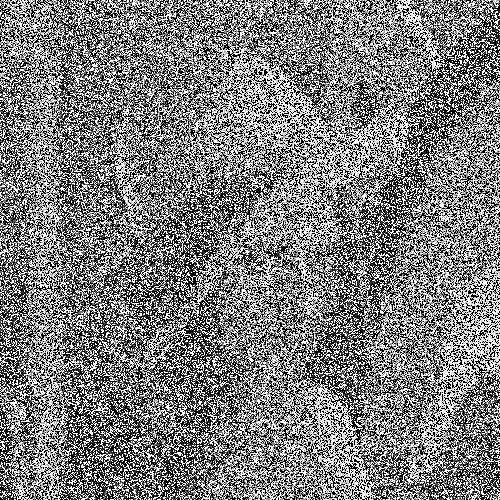}
\caption{}
\end{subfigure}
\begin{subfigure}[b]{.19\textwidth}
\centering
\includegraphics[scale=.17]{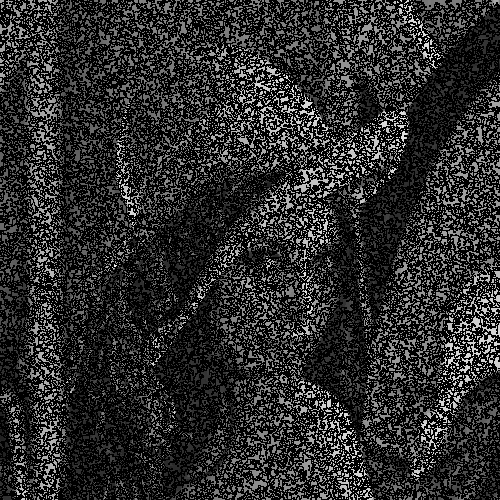}
\caption{}
\end{subfigure}
\begin{subfigure}[b]{.19\textwidth}
\centering
\includegraphics[scale=.17]{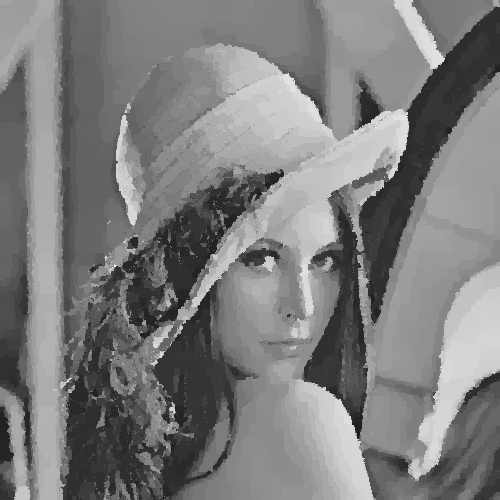}
\caption{}
\end{subfigure}\\
\begin{subfigure}[b]{.19\textwidth}
\centering
\includegraphics[scale=.17]{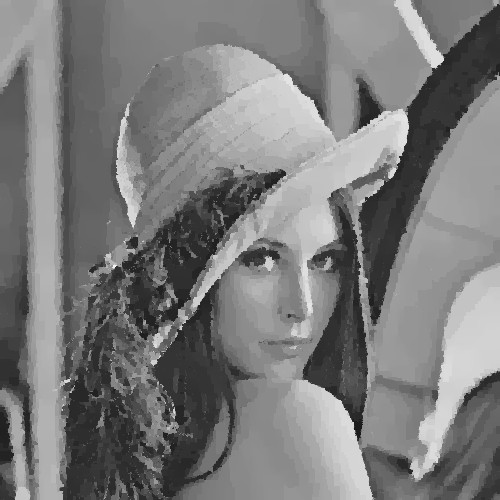}
\caption{}
\end{subfigure}
\begin{subfigure}[b]{.19\textwidth}
\centering
\includegraphics[scale=.17]{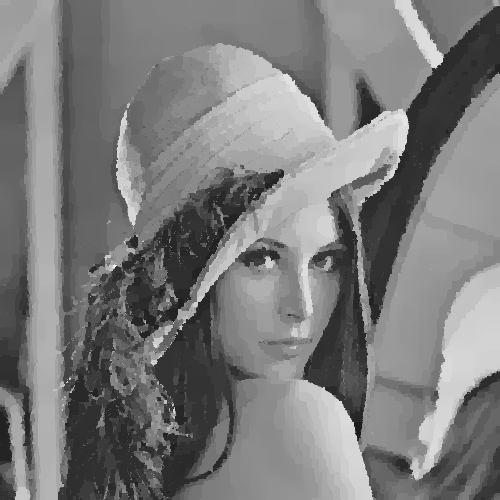}
\caption{}
\end{subfigure}
\begin{subfigure}[b]{.19\textwidth}
\centering
\includegraphics[scale=.17]{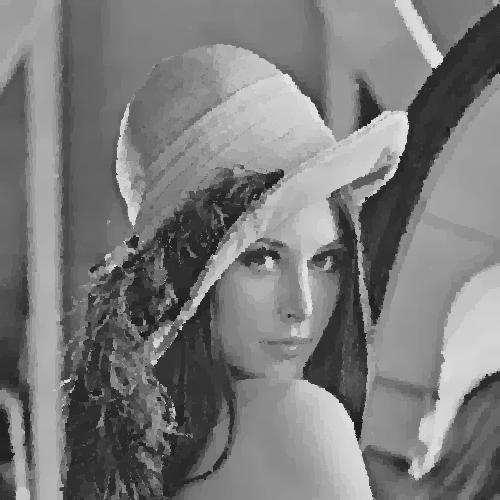}
\caption{}
\end{subfigure}
\begin{subfigure}[b]{.19\textwidth}
\centering
\includegraphics[scale=.17]{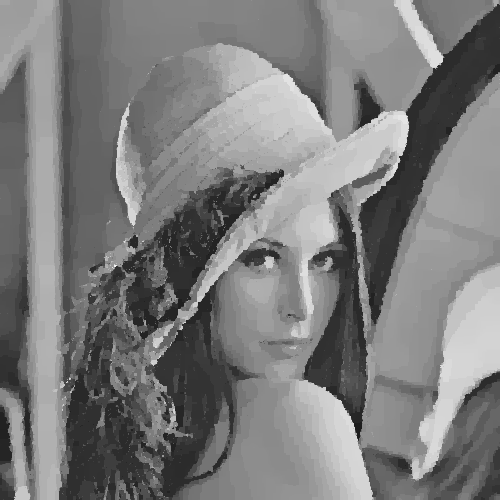}
\caption{}
\end{subfigure}
\vskip -0.35cm
\caption{The results of \textit{Lena} corrupted by 70\% salt and pepper noise, the numbers indicate PSNR and SSIM: (a) Ground truth, (b) Noisy images (7.00, 0.01), (c) ICM (7.61, 0.03), (d) Swap (26.15, 0.80), (e) BP-S (26.52, 0.81), (f) TRW-S (26.83, 0.81), (g) BP-M (26.83 , 0.82), (h) Expansion (\textbf{27.13}, 0.81)}
\label{fig:lenaResults}
\end{figure*}

For assessing the performance of the proposed graph cuts-based method for reduction of salt and pepper impulse noise, several test images (\textit{Boat}, \textit{Goldhill}, \textit{Lena} and \textit{Peppers}) with various levels (10-90\%) of impulse noise are considered. After detecting the pixels with min/max possible intensity values as corrupted pixels, a mask is created indicating the missing pixels to be estimated using the proposed approach. This mask is later used when defining the data and smoothness terms in the energy functional to be minimized. Figures \ref{fig:boatResults} (a) and \ref{fig:lenaResults} (a) show the initial \textit{Boat} and \textit{Lena} images, while Figures \ref{fig:boatResults} (b) and \ref{fig:lenaResults} (b) display the corresponding corrupted images by 50\% and 70\% salt and pepper impulse noise, respectively. 

Since the problem is formulated as a MRF model with smoothness priors, various techniques that are introduced for optimizing such problems are used for comparison, namely: Iterated Conditional Modes (ICM) \cite{besag1986statistical}, graph cuts with swap moves (Swap) \cite{boykov2004experimental,kolmogorov2004energy}, sequential and max-product loopy belief propagation (BP-S and BP-M) \cite{felzenszwalb2006efficient,freeman2000learning,tappen2003comparison} and sequential tree-reweighted message passing (TRW-S) \cite{wainwright2005map,kolmogorov2006convergent}. 

ICM as a deterministic \textit{greedy} strategy tries to find a local minimum. Starting from an estimate of the labeling and in an iterative manner, for each pixel, the label that gives the largest decrease of the energy function is chosen \cite{besag1986statistical}. Even though the method converges very fast in comparison to the other methods, the performance is very sensitive to the initial estimate. In graph cuts with swap moves, similar to the graph cuts with $\alpha$-expansion moves that is discussed in great details in previous sections, a repeating computation of the global minimum of a binary labeling problem is done which converges to a strong local minimum. Unlike the $\alpha$-expansion move, for a pair of labels $\alpha$ and $\beta$, the swap move considers a subset of pixels with label $\alpha$ and assigns them the label of $\beta$ and vice versa. This is done until there is no other swap move for any pair of labels that produces lower energy. 

Belief Propagation (BP) is a technique for exact inference of marginal probabilities for singly connected distributions \cite{barber2012bayesian}. However, the problems usually raised in the context of image processing and computer vision involve not singly connected but rather multiply connected (loopy) graphs. The variants of BP for loopy graphs, which are referred to as loopy belief propagation (LBP) techniques, have found significant attentions in the image processing and computer vision community \cite{felzenszwalb2006efficient,freeman2000learning,tappen2003comparison}. Here, the max-product LBP (BP-M) and sequential LBP (BP-S) are considered for comparison. The main difference between the two techniques is in the schedule of passing messages between image grid nodes. Similar to BP, TRW is considered as a message passing algorithm. However, the message update for an edge is weighted. The original TRW is shown to not necessarily converge \cite{wainwright2005map}, but the improved version, TRW-S, guaranties that the lower bound estimate of the energy never decreases \cite{kolmogorov2006convergent}. Here, the TRW-S as suggested by \cite{szeliski2008comparative} is used. The reader is referred to the above-mentioned references and the references therein for a more comprehensive discussion on the implementation aspects and properties of these methods\footnote{http://vision.middlebury.edu/MRF/}. 

Tables \ref{table:boatTable}-\ref{table:peppersTable} show the results of comparison between the proposed method and the other techniques mentioned above for salt and pepper noise reduction of the four sample images (\textit{Boat}, \textit{Goldhill}, \textit{Lena} and \textit{Peppers}) using PSNR and SSIM as quantitative metrics. The level of noise is changed between 10-90\%. The first row in each table represents the initial PSNR and SSIM of the noisy image while the rest are the results of ICM, graph cuts with swap moves, BP-S, BP-M, TRW-S and the proposed graph cuts based on $\alpha$-expansion moves, respectively. As can be seen from the tables, the PSNR and SSIM are decreased drastically in comparison to the original image even when having 10\% of corrupted pixels. The effect is worsened when having higher levels of noise. ICM manages to decrease the effects of noise greatly in comparison to the initial state, however, the final results still suffer from the residual missing pixels. This is the result of the principal strategy that is taken by the ICM which tries to find the local minimum of the energy function near the initial estimation of the labeling. Therefore, given a poor initial estimation, the final result of ICM is not very satisfying. This can be seen when we increase the level of noise, in which not only ICM cannot recover the image, the metrics are even lower than the initial state.

\begin{table*}[hbtp]
\footnotesize
\caption{PSNR and SSIM comparisons for various levels of noise (10-90\%) for \textit{Boat}} 
\centering 
\vskip -0.35cm
\resizebox{\textwidth}{!}{\begin{tabular}{c | c c c c c c c c c} 
\hline 
 & 10\% & 20\% & 30\% & 40\% & 50\% & 60\% & 70\% & 80\% & 90\% \\ [0.5ex] 
Method & PSNR  ,   SSIM & PSNR  ,   SSIM & PSNR  ,   SSIM & PSNR  ,   SSIM & PSNR  ,   SSIM & PSNR  ,  SSIM & PSNR  ,   SSIM & PSNR  ,   SSIM & PSNR  ,   SSIM\\
\hline \hline 
Initial& 15.36  , 0.21& 12.33  , 0.10& 10.56  , 0.06& 9.32  , 0.04& 8.35  , 0.03& 7.56  , 0.02& 6.88  , 0.01& 6.30  , 0.01& 5.79  , 0.00\\ 
ICM& 27.26  , 0.91& 20.86  , 0.75& 16.37  , 0.50& 12.85  , 0.27& 10.05  , 0.12& 8.10  , 0.07& 6.78  , 0.04& 5.93  , 0.03& 5.32  , 0.01\\ 
Swap& 34.91  , 0.95& 32.39  , 0.94& 30.57  , 0.92& 28.88  , 0.89& 27.45  , 0.86& 25.94  , 0.83& 24.26  , 0.77& 22.07  , 0.70& 17.81  , 0.58\\ 
BP-S& 35.82  , 0.96& 33.04  , 0.94& 31.04  , 0.92& 29.49  , 0.90& 28.08  , 0.87& 26.52  , 0.84& 24.84  , 0.79& 22.74  , 0.72& 19.87  , 0.62\\
TRW-S& 35.83  , 0.96& 33.10  , 0.94& 31.13  , 0.92& 29.58  , 0.90& 28.08  , 0.87& 26.61  , 0.84& 24.99  , 0.79& 22.92  , 0.72& 20.15  , 0.62\\  
BP-M& 35.86  , 0.96& 33.10  , 0.94& 31.04  , 0.92& 29.47  , 0.90& 28.08  , 0.88& 26.60  , 0.84& \textbf{25.15}  , 0.80& \textbf{23.23}  , 0.74& \textbf{21.37}  , 0.65\\ 
Expansion& \textbf{36.35}  , 0.96& \textbf{33.60}  , 0.94& \textbf{31.81}  , 0.93& \textbf{29.94}  , 0.90& \textbf{28.38}  , 0.88& \textbf{26.81}  , 0.84& 25.06  , 0.79& 22.93  , 0.72& 20.26  , 0.62\\ 
 [1ex] 
\hline 
\end{tabular}}
\label{table:boatTable} 
\end{table*}

\begin{table*}[hbtp]
\footnotesize
\caption{PSNR and SSIM comparisons for various levels of noise (10-90\%) for \textit{Goldhill}} 
\centering 
\vskip -0.35cm
\resizebox{\textwidth}{!}{\begin{tabular}{c | c c c c c c c c c} 
\hline 
 & 10\% & 20\% & 30\% & 40\% & 50\% & 60\% & 70\% & 80\% & 90\% \\ [0.5ex] 
Method & PSNR  ,   SSIM & PSNR  ,   SSIM & PSNR  ,   SSIM & PSNR  ,   SSIM & PSNR  ,   SSIM & PSNR  ,  SSIM & PSNR  ,   SSIM & PSNR  ,   SSIM & PSNR  ,   SSIM\\
\hline \hline 
Initial& 15.41  , 0.20& 12.38  , 0.09& 10.60  , 0.05& 9.35  , 0.03& 8.37  , 0.02& 7.59  , 0.01& 6.93  , 0.01& 6.33  , 0.00& 5.83  , 0.00\\ 
ICM& 28.34  , 0.89& 21.96  , 0.71& 17.48  , 0.46& 14.10  , 0.23& 11.41  , 0.11& 9.52  , 0.06& 8.28  , 0.04& 7.43  , 0.02& 6.84  , 0.01\\ 
Swap& 36.18  , 0.94& 33.74  , 0.92& 32.06  , 0.90& 30.59  , 0.87& 29.26  , 0.83& 27.91  , 0.79& 26.32  , 0.73& 24.13  , 0.64& 20.39  , 0.50\\ 
BP-S& 36.50  , 0.94& 34.15  , 0.92& 32.45  , 0.90& 30.95  , 0.87& 29.59  , 0.84& 28.19  , 0.80& 26.77  , 0.75& 24.99  , 0.67& 22.28  , 0.56\\
TRW-S& 36.43  , 0.94& 34.18  , 0.92& 32.52  , 0.90& 31.04  , 0.87& 29.68  , 0.84& 28.33  , 0.80& 27.02  , 0.75& 25.35  , 0.67& 22.74  , 0.57\\  
BP-M& 36.56  , 0.94& 34.20  , 0.92& 32.47  , 0.90& 30.97  , 0.88& 29.63  , 0.84& 28.26  , 0.80& 27.06  , 0.76& \textbf{25.62}  , 0.69& \textbf{23.70}  , 0.60\\ 
Expansion& \textbf{36.68}  , 0.94& \textbf{34.39}  , 0.92& \textbf{32.78}  , 0.90& \textbf{31.31}  , 0.87& \textbf{29.82}  , 0.84& \textbf{28.44}  , 0.80& \textbf{27.10}  , 0.75& 25.34  , 0.67& 22.37  , 0.56
\\ [1ex] 
\hline 
\end{tabular}}
\label{table:goldhillTable} 
\end{table*}

\begin{table*}[hbtp]
\footnotesize
\caption{PSNR and SSIM comparisons for various levels of noise (10-90\%) for \textit{Lena}} 
\centering 
\vskip -0.35cm
\resizebox{\textwidth}{!}{\begin{tabular}{c | c c c c c c c c c} 
\hline 
 & 10\% & 20\% & 30\% & 40\% & 50\% & 60\% & 70\% & 80\% & 90\% \\ [0.5ex] 
Method & PSNR  ,   SSIM & PSNR  ,   SSIM & PSNR  ,   SSIM & PSNR  ,   SSIM & PSNR  ,   SSIM & PSNR  ,  SSIM & PSNR  ,   SSIM & PSNR  ,   SSIM & PSNR  ,   SSIM\\
\hline \hline 
Initial& 15.50  , 0.17& 12.45  , 0.08& 10.68  , 0.05& 9.41  , 0.03& 8.46  , 0.02& 7.65  , 0.01& 7.00  , 0.01& 6.42  , 0.00& 5.89  , 0.00\\ 
ICM& 29.05  , 0.91& 22.31  , 0.76& 17.63  , 0.51& 14.03  , 0.25& 11.17  , 0.10& 8.94  , 0.05& 7.61  , 0.03& 6.75  , 0.02& 6.13  , 0.01\\ 
Swap& 36.91  , 0.95& 34.08  , 0.93& 32.33  , 0.91& 30.87  , 0.90& 29.48  , 0.87& 28.19  , 0.84& 26.15  , 0.80& 23.38  , 0.73& 19.40  , 0.62\\ 
BP-S& 37.14  , 0.94& 34.44  , 0.93& 32.72  , 0.92& 31.13  , 0.90& 29.81  , 0.88& 28.41  , 0.85& 26.52  , 0.81& 24.40  , 0.75& 20.99  , 0.66\\
TRW-S& 37.09  , 0.94& 34.44  , 0.93& 32.78  , 0.91& 31.39  , 0.90& 29.94  , 0.88& 28.62  , 0.85& 26.83  , 0.81& 24.78  , 0.76& 21.04  , 0.66\\  
BP-M& 37.19  , 0.95& 34.45  , 0.93& 32.70  , 0.92& 31.18  , 0.90& 29.90  , 0.88& 28.50  , 0.85& 26.83  , 0.82& \textbf{24.98}  , 0.77& \textbf{22.64}  , 0.69\\ 
Expansion& \textbf{37.44}  , 0.94& \textbf{35.00}  , 0.93& \textbf{33.13}  , 0.92& \textbf{31.62}  , 0.90& \textbf{30.27}  , 0.88& \textbf{28.66}  , 0.85& \textbf{27.13}  , 0.81& 24.83  , 0.76& 21.23  , 0.66\\ 
 [1ex] 
\hline 
\end{tabular}}
\label{table:lenaTable} 
\end{table*}

\begin{table*}[hbtp]
\footnotesize
\caption{PSNR and SSIM comparisons for various levels of noise (10-90\%) for \textit{Peppers}} 
\centering 
\vskip -0.35cm
\resizebox{\textwidth}{!}{\begin{tabular}{c | c c c c c c c c c} 
\hline 
 & 10\% & 20\% & 30\% & 40\% & 50\% & 60\% & 70\% & 80\% & 90\% \\ [0.5ex] 
Method & PSNR  ,   SSIM & PSNR  ,   SSIM & PSNR  ,   SSIM & PSNR  ,   SSIM & PSNR  ,   SSIM & PSNR  ,  SSIM & PSNR  ,   SSIM & PSNR  ,   SSIM & PSNR  ,   SSIM\\
\hline \hline 
Initial& 15.31  , 0.16& 12.26  , 0.07& 10.49  , 0.04& 9.26  , 0.03& 8.31  , 0.02& 7.50  , 0.01& 6.84  , 0.01& 6.25  , 0.00& 5.76  , 0.00\\ 
ICM& 30.01  , 0.91& 23.28  , 0.78& 18.16  , 0.53& 14.29  , 0.28& 11.34  , 0.14& 9.22  , 0.07& 7.89  , 0.05& 7.03  , 0.03& 6.42  , 0.02\\ 
Swap& 36.32  , 0.93& 33.98  , 0.91& 32.50  , 0.90& 31.15  , 0.88& 29.96  , 0.85& 28.40  , 0.83& 26.49  , 0.79& 23.84  , 0.73& 18.97  , 0.61\\ 
BP-S& 36.40  , 0.93& 34.45  , 0.91& 32.77  , 0.89& 31.45  , 0.88& 30.17  , 0.86& 28.76  , 0.83& 26.85  , 0.79& 24.85  , 0.74& 20.51  , 0.64\\
TRW-S& 36.33  , 0.92& 34.47  , 0.91& 32.82  , 0.89& 31.59  , 0.88& 30.29  , 0.86& 28.88  , 0.83& 27.12  , 0.80& 25.12  , 0.75& 21.36  , 0.66\\  
BP-M& 36.46  , 0.93& 34.45  , 0.91& 32.78  , 0.90& 31.48  , 0.88& 30.13  , 0.86& 28.86  , 0.83& \textbf{27.27}  , 0.80& \textbf{25.36}  , 0.75& \textbf{22.92}  , 0.68\\ 
Expansion& \textbf{36.70}  , 0.92& \textbf{34.89}  , 0.91& \textbf{33.28}  , 0.89& \textbf{31.88}  , 0.88& \textbf{30.68}  , 0.86& \textbf{29.14}  , 0.83& 27.21  , 0.80& 25.08  , 0.74& 21.60  , 0.65\\ 
 [1ex] 
\hline 
\end{tabular}}
\label{table:peppersTable} 
\end{table*}

In contrast, the rest of the techniques try to find a strong minimum for the MRF based energy minimization problem and manage to achieve this, of course at the price of more computational time. Comparing the results of the techniques, the proposed graph cuts with $\alpha$-expansion moves approach outperforms the rest of the techniques in most cases in terms of PSNR. This is the case for all of the test images. However, for noise levels of more than 70\%, the performance of the proposed approach decreases, although it remains very close to the best achieved performance. Overall, the performance of the graph cuts with $\alpha$-expansion moves method is shown to be superior to the other message passing techniques such as BP-S, BP-M and TRW-S for the problem of salt and pepper noise reduction. In terms of SSIM, the performances of all techniques, except the ICM, are in the same order with minimal variations. However, BP-M shows superiority when having salt and pepper noise levels higher than 70\% both in terms of PSNR and SSIM. Figures \ref{fig:boatResults} and \ref{table:lenaTable} show the results of the proposed graph cuts based technique as well as the results of the other techniques described above for \textit{Boat} with 50\% noise and \textit{Lena} with 70\% noise. As expected, ICM falls into a local minimum very close to the initial labeling estimate which results in having residual missing pixels. On a qualitative level, the results of the other techniques are very similar. However, the metrics provide a more quantitative measure for the accuracy of the techniques. 

\begin{figure*}[hbtp]
\centering
\begin{subfigure}[b]{.19\textwidth}
\centering
\includegraphics[scale=.17]{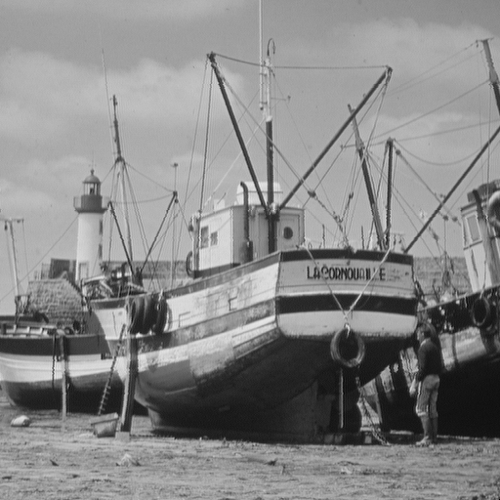}
\caption{}
\end{subfigure}
\begin{subfigure}[b]{.19\textwidth}
\centering
\includegraphics[scale=.17]{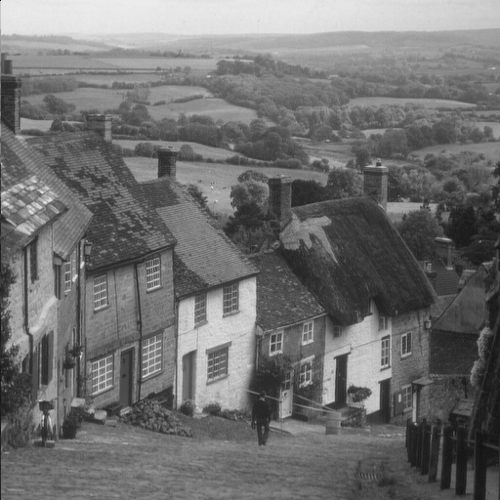}
\caption{}
\end{subfigure}
\begin{subfigure}[b]{.19\textwidth}
\centering
\includegraphics[scale=.17]{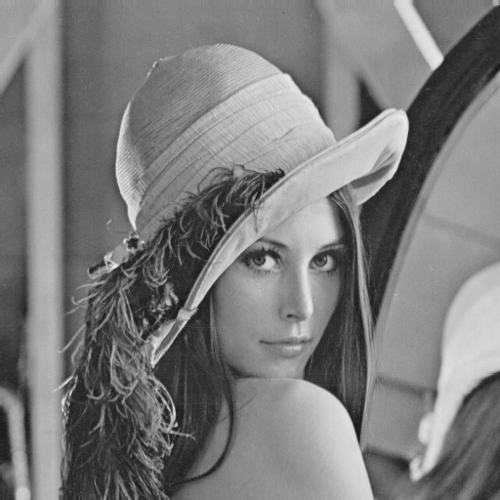}
\caption{}
\end{subfigure}
\begin{subfigure}[b]{.19\textwidth}
\centering
\includegraphics[scale=.17]{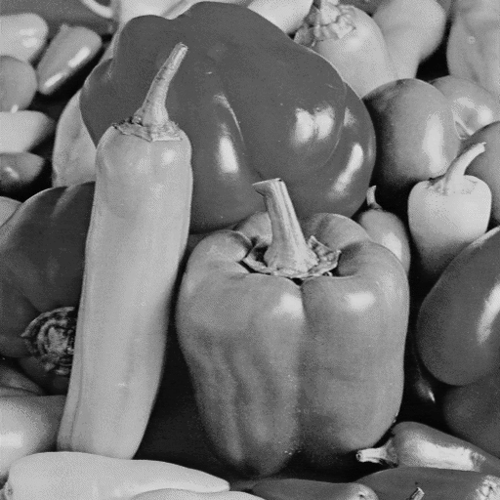}
\caption{}
\end{subfigure}\\
\begin{subfigure}[b]{.19\textwidth}
\centering
\includegraphics[scale=.17]{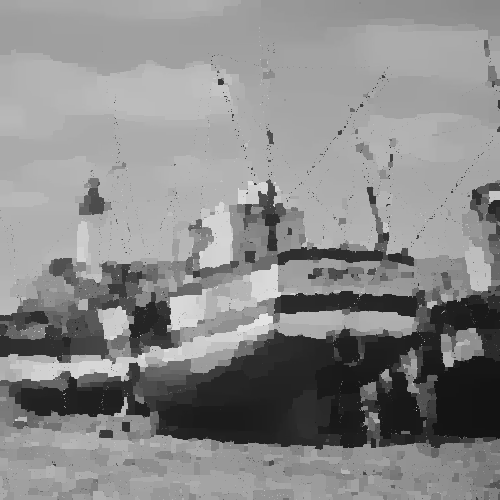}
\caption{}
\end{subfigure}
\begin{subfigure}[b]{.19\textwidth}
\centering
\includegraphics[scale=.17]{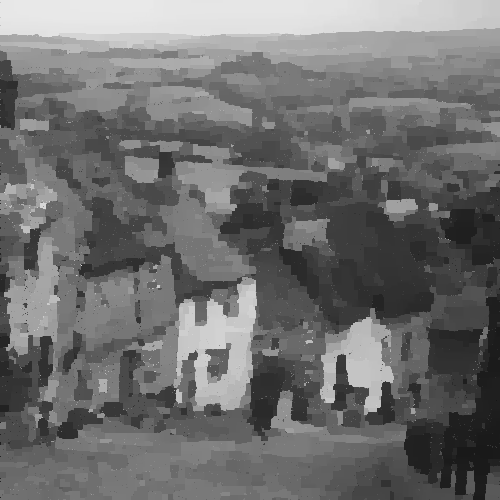}
\caption{}
\end{subfigure}
\begin{subfigure}[b]{.19\textwidth}
\centering
\includegraphics[scale=.17]{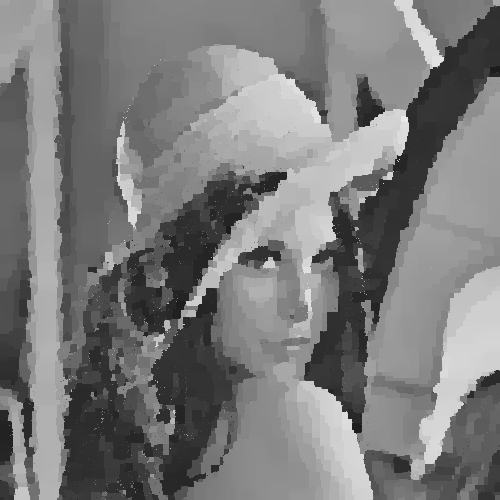}
\caption{}
\end{subfigure}
\begin{subfigure}[b]{.19\textwidth}
\centering
\includegraphics[scale=.17]{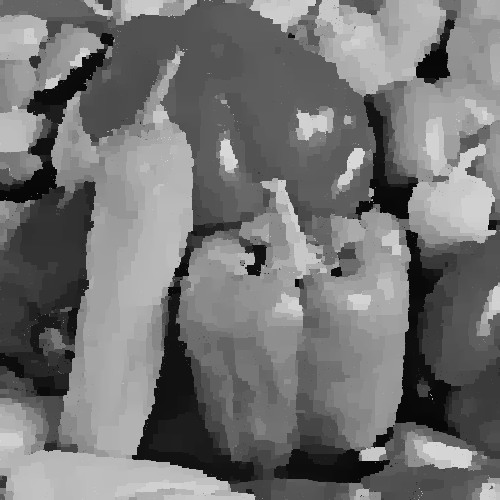}
\caption{}
\end{subfigure}\\
\begin{subfigure}[b]{.19\textwidth}
\centering
\includegraphics[scale=.17]{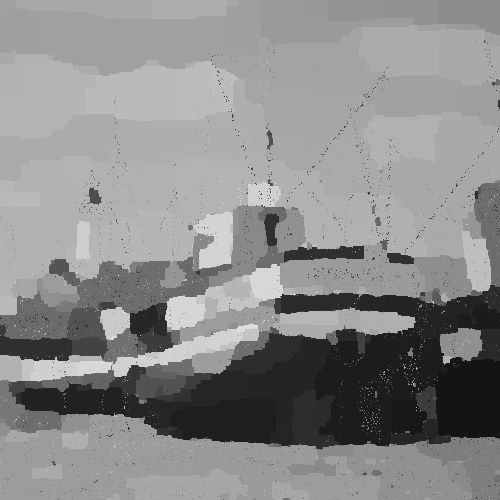}
\caption{}
\end{subfigure}
\begin{subfigure}[b]{.19\textwidth}
\centering
\includegraphics[scale=.17]{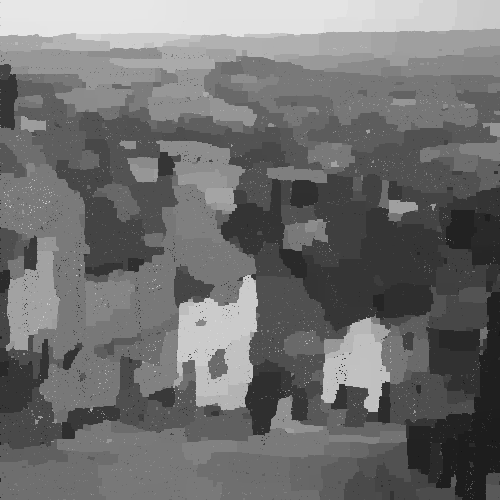}
\caption{}
\end{subfigure}
\begin{subfigure}[b]{.19\textwidth}
\centering
\includegraphics[scale=.17]{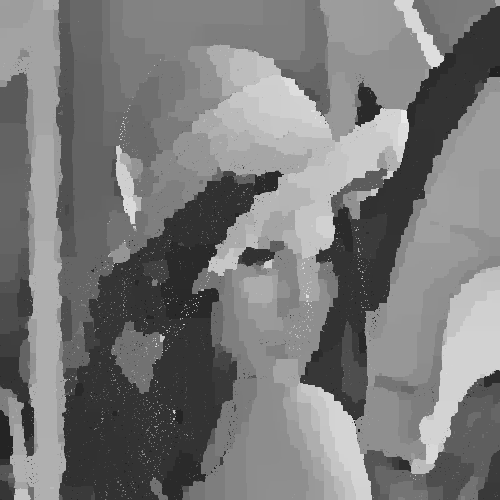}
\caption{}
\end{subfigure}
\begin{subfigure}[b]{.19\textwidth}
\centering
\includegraphics[scale=.17]{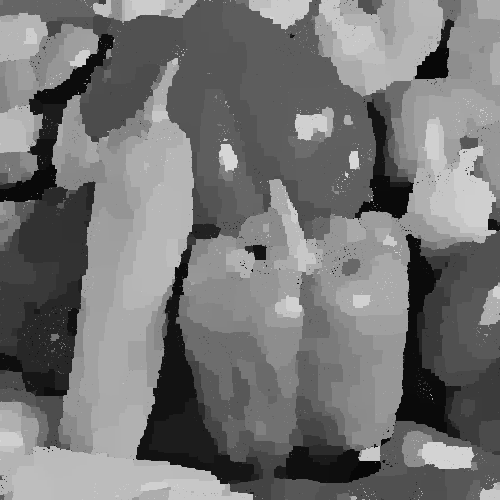}
\caption{}
\end{subfigure}
\vskip -0.35cm
\caption{The results of test images corrupted by 90\% salt and pepper noise for comparison between the BP-M and the proposed graph cuts with expansion moves: (a-d) Ground truth images of \textit{Boat}, \textit{Goldhill}, \textit{Lena} and \textit{Peppers}, (e-h) Denoised results using BP-M, (i-l) Denoised results using the proposed method utilizing graph cuts with expansion moves.}
\label{fig:highnoiseResults}
\end{figure*}

Figure \ref{fig:highnoiseResults} shows the visual comparison between the proposed graph cuts with $\alpha$-expansion moves and that of BP-M for when the level of noise is set to 90\%. It should be noted that the graph cuts with expansion moves is designed to account for \textit{large moves} in the constructed graph to achieve the strong minimum for the MRF model. Due to high level of noise, which is equivalent to a very sparse distribution of known pixels, the method tends to generate a very piece-wise smooth denoised image. On the other hand, the BP-M which is based on message passing among local neighbors can recover a more accurate image, both in terms of PSNR and SSIM when having very high density salt and pepper noise. 

\begin{figure}[!t]
\centering
\includegraphics[scale=.5]{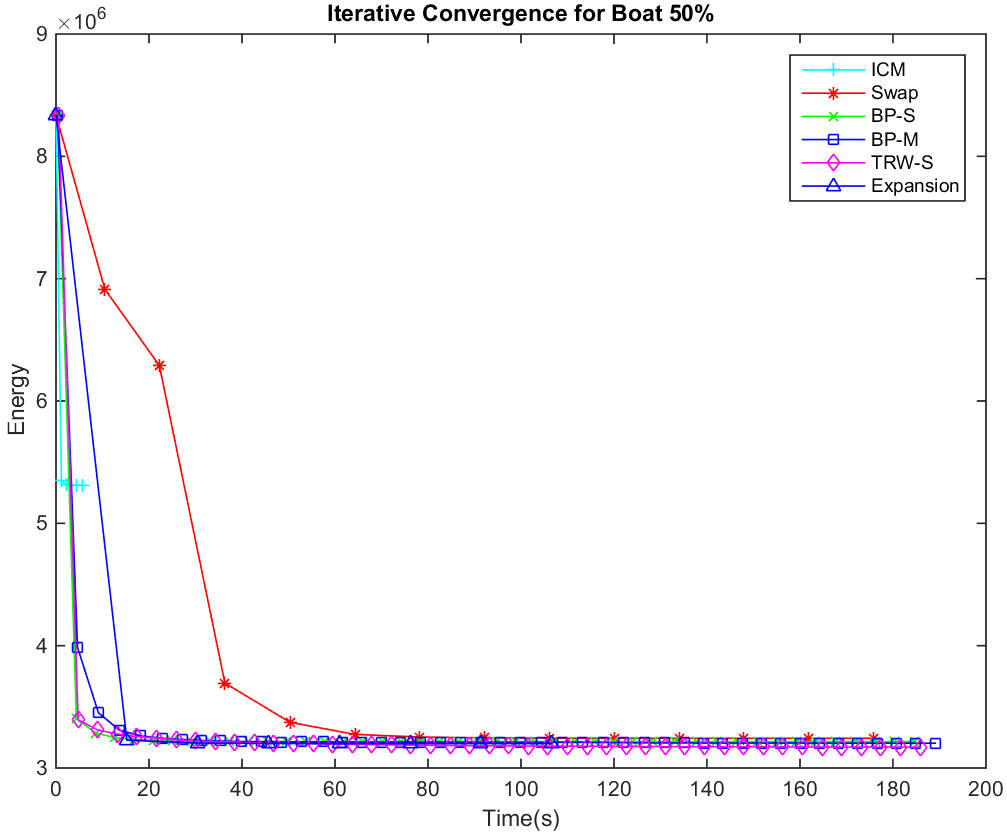}
\vskip -0.35cm
\caption{The iterative convergence curves for \textit{Boat} with 50\% of corrupted pixels}
\label{fig:iterativeConv}
\end{figure}

For completeness of the comparisons, computational time can also be considered for comparison between different methods. Figure \ref{fig:iterativeConv} shows the iterative convergence of various techniques for the \textit{Boat} with 50\% corrupted pixels. As is obvious, ICM is very fast but with the optimized energy much larger than what can be achieved with other techniques. On the other hand, the proposed method and the other techniques reach the same minimum approximately, although at different times. A more close investigation reveals that while the computational time for each iteration of graph cuts with expansion moves is more than the others, the solution becomes stable in fewer iterations which results in less overall computational time. 

\subsection{Possible Enhancements \& Future Directions}

\subsubsection{Neighborhood System: First, Second, ... Order?}
For the current study, as mentioned before, a first order 4-connected neighborhood system is used for the purpose of impulse noise reduction. As one can argue, having more neighbors may result in more accurate results since additional information can be gathered in order to reconstruct the image, especially in the edge regions. This can be even moved further with higher order neighborhood systems incorporating information from pixels that are not in the direct neighborhood of the pixel of interest. This is left to future research. 

\subsubsection{Super-Resolution from a Single Image}
\begin{figure}[t]
\centering
\begin{subfigure}[b]{.15\textwidth}
\centering
\includegraphics[scale=.155]{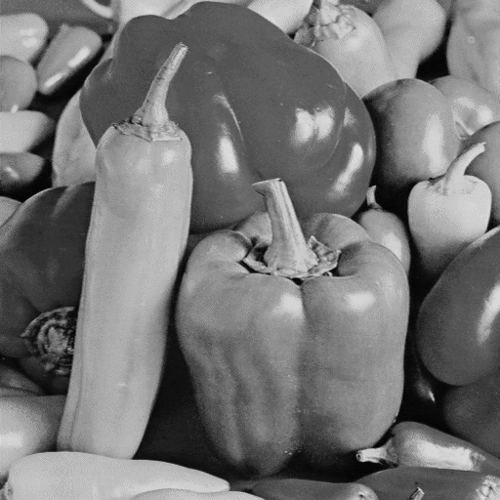}
\caption{}
\end{subfigure}
\begin{subfigure}[b]{.15\textwidth}
\centering
\includegraphics[scale=.155]{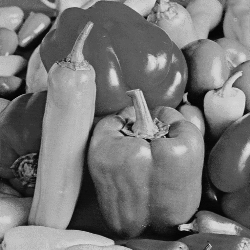}
\caption{}
\end{subfigure}
\begin{subfigure}[b]{.15\textwidth}
\centering
\includegraphics[scale=.155]{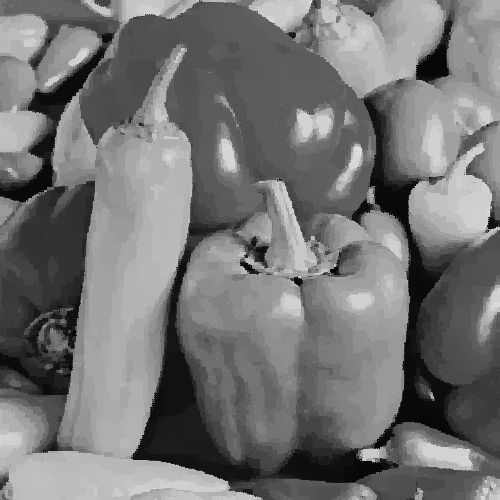}
\caption{}
\end{subfigure}\\
\begin{subfigure}[b]{.15\textwidth}
\centering
\includegraphics[scale=.155]{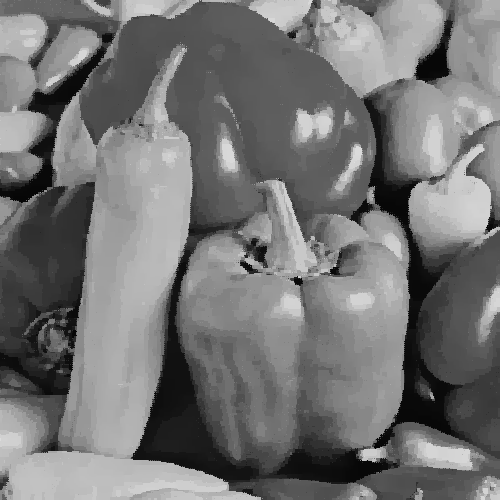}
\caption{}
\end{subfigure}
\begin{subfigure}[b]{.15\textwidth}
\centering
\includegraphics[scale=.155]{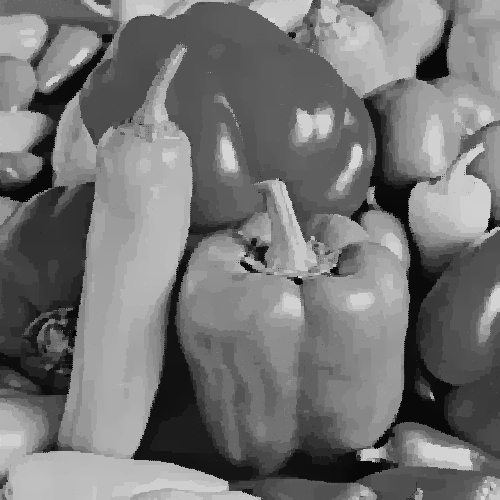}
\caption{}
\end{subfigure}
\begin{subfigure}[b]{.15\textwidth}
\centering
\includegraphics[scale=.155]{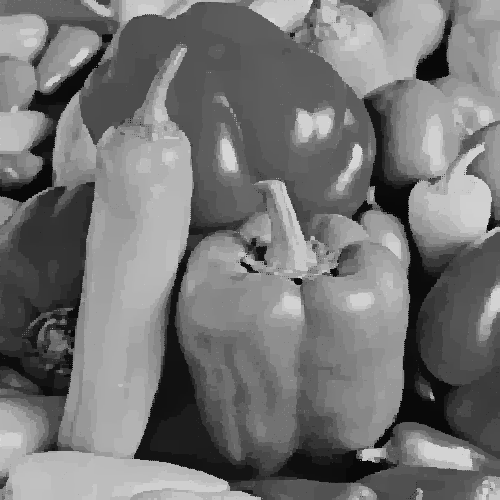}
\caption{}
\end{subfigure}
\vskip -0.35cm
\caption{The results of $\times2$ interpolation for \textit{Peppers} and the comparison metrics (PSNR, SSIM): (a) Ground truth, (b) Downsampled image, (c) BP-S (28.20, 0.81), (d) BP-M (28.04, 0.81), (e) TRW-S (28.46, 0.81), (f) Expansion (28.58, 0.81)}
\label{fig:interpSample}
\end{figure}

The problem of image interpolation/super-resolution can be considered as a special case of the framework used here. As mentioned before, here, at first the noisy pixels are detected and used to generate a mask indicating the corrupted pixels which was later used for determining the data and smoothness components of the energy functional to be minimized. Assume the general case of up-sampling a low-resolution image by a factor of 2, which is generally considered by the literature in this area \cite{baghaie2015structure}. This can be implemented in the same manner, although the implications regarding the appropriate parameters chosen for the energy minimization problem can be different. For this, at first, the pixels from the original low-resolution image are placed in their appropriate locations in the high-resolution grid. In this case, the high-resolution image is missing about 75\% of the pixels which can be estimated by the same framework that is used for salt and pepper noise reduction. The main difference is in the pattern of missing pixels, which unlike the impulse noise case where the distribution is random, here, we have a uniform grid of missing values that need to be approximated. It should be mentioned that this needs more investigation on the implications of such high amount of missing pixels since, as discussed before, excessive amount of corrupted/missing pixels may impact the accuracy of the reconstructed results greatly. Figure \ref{fig:interpSample} shows an example of single image super-resolution using the graph cuts with $\alpha$-expansion moves as well as other energy minimization techniques.

\section{Conclusion}
\label{S:4}
Noise reduction is a common practice in many image processing problems as a pre-processing step, with many different techniques in the literature for various types of noise. In the current work, the problem of impulse noise reduction from images with a focus on fixed-valued impulse noise, also known as salt and pepper noise, is considered. Having an image corrupted by salt and pepper noise, at first, the corrupted pixels are detected by a min/max detection approach. This results in a mask indicating the locations of the corrupted pixels which is later used for reconstruction of the noise-free image. The reconstruction process is formulated as an inpaiting problem by considering a Markov Random Field (MRF) model with smoothness priors. The resulting energy functional is then minimized by use of graph cuts with $\alpha$-expansion moves. For assessing the performance of the proposed method, several standard test images with various levels of salt and pepper noise ranging from 10 to 90\% are considered. Using PSNR and SSIM as comparison metrics, the performance of the proposed method is compared with several state-of-the-art MRF energy minimization techniques for the same formulation. While the comparison between SSIM values is representative of minimal differences between various techniques, the proposed method outperforms the other techniques with significant margin in terms of PSNR. Moreover, the computational time is less than the rest of the methods considered here. Further improvements can be achieved by using more sophisticated neighborhood systems, which is left to future research.








%

\bibliographystyle{IEEEtran}
\bibliography{sample}




\end{document}